\documentclass[nohyperref]{article}

\usepackage{amsmath}
\usepackage{microtype}
\usepackage{graphicx}
\usepackage{subfigure}
\usepackage{booktabs} 
\usepackage{enumitem}
\usepackage{multirow}
\usepackage{ragged2e}

\usepackage{hyperref}
\usepackage{accents}

\usepackage[accepted]{icml2022}

\usepackage{savesym}
\usepackage{amssymb}
\usepackage{mathtools}
\usepackage{amsthm}

\usepackage[capitalize,noabbrev]{cleveref}

\theoremstyle{plain}

\theoremstyle{definition}

\theoremstyle{remark}

\usepackage[textsize=tiny]{todonotes}

\icmltitlerunning{FairGen: Fair Synthetic Data Generation}

\begin{document}
\twocolumn[
\icmltitle{FairGen: Fair Synthetic Data Generation }



\icmlsetsymbol{equal}{*}

\begin{icmlauthorlist}
\icmlauthor{Himanshu Chaudhary}{equal,comp}
\icmlauthor{Bhushan Chaudhari}{equal,comp}
\icmlauthor{Aakash Agarwal}{comp2}
\icmlauthor{Kamna Meena}{comp}
\icmlauthor{Tanmoy Bhowmik}{comp3}
\end{icmlauthorlist}

\icmlaffiliation{comp}{Mastercard, India}
\icmlaffiliation{comp2}{Credgenics, India}
\icmlaffiliation{comp3}{GoTo Group, India}


\icmlcorrespondingauthor{Himanshu Chaudhary}{himanshu.chaudhary@mastercard.com}
\icmlcorrespondingauthor{Bhushan Chaudhari}{bhushan.chaudhari@mastercard.com}
\icmlcorrespondingauthor{Aakash Agarwal}{aakash.agarwal1307@gmail.com}
\icmlcorrespondingauthor{Kamna Meena}{mkamna14@gmail.com}
\icmlcorrespondingauthor{Tanmoy Bhowmik}{tantanmoy@gmail.com}

\icmlkeywords{Machine Learning, Fairness, GAN, Synthetic Data}

\vskip 0.3in
]



\hbadness=99999  
\printAffiliationsAndNotice{\icmlEqualContribution}
\begin{abstract}
With the rising adoption of Machine Learning across the domains like banking, pharmaceutical, ed-tech, etc, it has become utmost important to adopt responsible AI methods to ensure models are not unfairly discriminating against any group. Given the lack of clean training data, generative adversarial techniques are preferred to generate synthetic data with several state-of-the-art architectures readily available across various domains from unstructured data such as text, images to structured datasets modelling fraud detection and many more. These techniques overcome several challenges such as class imbalance, limited training data, restricted access to data due to privacy issues. Existing work focusing on generating fair data either works for a certain GAN architecture or is very difficult to tune across the GANs. In this paper, we propose a pipeline to generate fairer synthetic data independent of the GAN architecture. The proposed paper utilizes a pre-processing algorithm to identify and remove bias inducing samples. In particular, we claim that while generating synthetic data most GANs amplify bias present in the training data but by removing these bias inducing samples, GANs essentially focuses more on real informative samples. Our experimental evaluation on two open-source datasets demonstrates how the proposed pipeline is generating fair data along with improved performance in some cases.
\end{abstract}

\section{Introduction}
\label{submission}

Machine learning algorithms are being used ubiquitously in the current world ranging from recommendation systems, fraud detection, facial recognition to Autonomous driving. These algorithms uses historical data to automate decision making. However, these machine learning models have unintended bias towards protected group as demonstrated by \cite{Calders2013}. Biasness comes in the model due to the training data being biased itself. Training data can have unfairness based on personal protected attributes such as gender race, religion, etc. and this unfairness in data doesn’t limit to only real world data but it also get propagated to models which are built on such biased data.
\begin{figure*}[thpb]
    \centering
    \includegraphics[width=\textwidth,height=4.5cm]{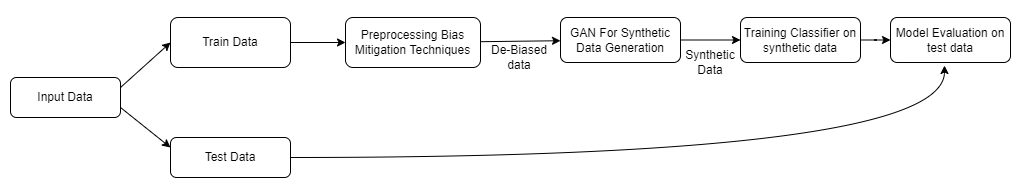}
    \caption{Data Flow}
    \label{fig:my_image}
\end{figure*}
Deep Learning models require huge training data to be accurate. But due to limited training data, class imbalance issues, Synthetic data have gained prominence as a replacement to get additional high quality real world training data. Generative Adversarial Networks (GANs) are one of the most used popular way to generate synthetic data \cite{GAN}. GANs are generative models that has two components: generator and discriminator. Model is trained in such a way that generator fools discriminator and generates synthetic data almost similar to real data. This generated high quality synthetic data can be used for different predictive analysis in case where real world data is limited
\cite{Edward2017}. Health and financial datasets are particularly ripe for a synthetic approach. These fields are highly restricted by privacy laws. Synthetic data can help researchers get the data they need without violating people’s privacy. For institutions in health, financial sectors they can not share their customers' data directly. However if they want to open source their data by adhering to all regulatory rules for privacy of their customers, then they can create synthetic data using real customer's data. In doing so, the real data will not get shared and the generated synthetic data has the same distributional properties as of real data and this can be used by researchers to create and do groundbreaking research. 

Existing GAN techniques generates biased data by amplifying the bias present in the training data.
\cite{Anubha} showed bias amplification across various GAN architectures including deferentially private generation schemes. This paper motivated us to find ways of reducing bias from the training data itself.

FairGAN \cite{FairGAN} and FairGAN+ \cite{FairGAN+} are GAN architectures that generates synthetic data that is bias free. FairGAN consists of two components: a generator G which generates the fake data conditioned on the protected attribute S; $P_G(x,y,s) = P_G(x,y|s)P_G(s)$ where $P_G(s) = P_{data}(s)$, and instead of one discriminator, it has two discriminators $D_1$ and $D_2$. $D_1$ is trained to differentiate the real data denoted by $P_{data}(x,y,s)$ from the generated fake data denoted by $P_G(x,y,s)$. And, for achieving fairness constraints, such as statistical parity \cite{Cynthia2012}, the training of $D_2$ is such that it emphasizes differentiation of the two types of synthetic (generated by the model) samples $P_G(x,y|s = 1)$ and $P_G(x,y|s = 0)$ indicating if the synthetic samples are from the privileged group or unprivileged group. By playing adversarial games with both the discriminators, the generator can generate high quality fake samples and does not encode any information supporting to predict the value of the protected attribute. But FairGAN is not helpful in case where classification-based fairness notions such as Average Odds Difference \cite{Rachel} or Equal Opportunity Difference \cite{Rachel} is important. FairGAN+ is an improved version of FairGAN which overcomes these limitations. It has an extra discriminator $D_3$ which is trained to distinguish values of the protected attribute from the predictions made by classifier. 
$D_3$ is trained to correctly predict S given a sample while the classifier aims to fool the discriminator $D_3$ and aims to make the prediction of samples not encode any information supporting to predict the value of the protected attribute. Once the prediction of the classifier cannot be used to predict the protected attribute S, the correlation between them is removed. But both of these architecture comes with limitations: For fairGan it needs to tune 3-deep learning models(1 generator and 2 discriminators) and for fairGan+ it needs to tune 4-deep learning models(1 generator and 3 discriminators) which makes the model even more complex and very hard to train. In addition to this, replicating this complex network to different other GAN architectures like CTGAN, Copula GAN is quite difficult.

Inspired by limitations of fairGAN and fairGAN+, we are proposing a pipeline to generate fairer synthetic data from any existing state-of-the-art GAN architectures by identifying and removing bias inducing samples in the training data using a preprocessing based bias mitigation method. K\% Removal Technique \cite{Aakash} is used to preprocess the training data and then the bias mitigated preprocessed data is utilized to generate synthetic data. We empirically found that the model developed on this synthetically generated fair data is fairer as well as perform better in some cases.

To summarize, in this paper, we make the following contributions :
    \begin{enumerate}[noitemsep,nolistsep]
        \item We propose a pipeline to generate fairer synthetic data from any existing state-of-the-art GAN architectures by identifying and removing bias inducing samples in the training data using a preprocessing based bias mitigation method. 
        \item Model developed on the synthetically generated data using the proposed pipeline is fairer than the model developed on the raw data.
        \item The proposed pipeline works seamlessly across the GAN architectures without adding any computationally expensive load on the architectures.
        \item Further, we comprehensively bias tested the pipeline for the intersectional bias testing along with standard fairness metrics.
    \end{enumerate}

\begin{center}
\begin{table*}
  \centering
  \begin{tabular}{|c| c| c| c| c|c|c| c| c|c|c|}
    \hline
   \multirow{2}{*}{GANs}  &  Preprocessing bias& \multicolumn{3}{|c|}{\textbf{Gender}} &
    \multicolumn{3}{|c|}{\textbf{Race}} & \multicolumn{3}{|c|}{\textbf{Intersectional}}\\
    \cline{3-11}
   &  mitigation technique & BCA & DPR & EOddR& BCA & DPR & EOddR& BCA & DPR & EOddR\\ [0.5ex] 
    \hline
      \multirow{5}{*}{CTGAN}& Raw & 0.59 & 0.89 & 0.97 & 0.63 & 0.83 &0.85 & 0.56& 0.75 &0.78 \\
      & \text{1 \% removal} & 0.56 & 0.91 & 0.94 & {\bf0.56} & {\bf0.94} &{\bf0.99} & 0.56 & 0.55 & 0.47 \\
      & \text{2 \% removal} & 0.59 & 0.87 & 0.91 & 0.58 & 0.93 & 0.74 & 0.53 & 0.42 & 0.57 \\
      & \text{3 \% removal} & {\bf0.63} & {\bf0.95} & {\bf0.99} & 0.56 & 0.99 & 0.94 &{\bf0.58} & {\bf0.80} & {\bf0.89} \\
      & \text{Augmentation} & 0.61 & 0.97 & 0.93 & 0.59 & 0.83 & 0.77 & - & - & -  \\
     \hline
     
     \multirow{5}{*}{Copula GAN}& Raw & 0.57 & 0.93 & 0.92 & 0.59 & 0.94 & 0.97 & 0.61 & 0.88 & 0.85 \\
      & \text{1 \% removal} & 0.53 & 0.94 & 0.97 & 0.60 & 0.99 & 0.95 & 0.61 & 0.95 & 0.83 \\
      & \text{2 \% removal} & {\bf0.56} & {\bf0.98} & {\bf0.94} & {\bf0.58} & {\bf0.97} & {\bf0.96} & 0.59 & 0.92 & 0.83 \\
      & \text{3 \% removal} & 0.50 & 0.93 & 0.94 & 0.48 & 0.93 & 0.82 & {\bf0.62} & {\bf0.89} & {\bf0.87} \\
      & \text{Augmentation} & 0.52 & 0.88 & 0.90 & 0.57 & 0.79 & 0.81 & - & - & - \\
     \hline
     
      & Raw & 0.61 & 0.93 & 0.95 & 0.63 & 0.85 & 0.93 & 0.62 & 0.83 & 0.73 \\
      & \text{1 \% removal} & 0.46 & 0.99 & 0.91 & {\bf0.60} & {\bf0.96} & {\bf0.99} & 0.63 & 0.81 & 0.83 \\
      Gaussian & \text{2 \% removal} & 0.52 & 0.99 & 0.95 & 0.61 & 0.84 & 0.90 & 0.62 & 0.83 & 0.79 \\
      Copula & \text{3 \% removal} &{\bf 0.54} & {\bf0.96} & {\bf0.96} & 0.51 & 0.90 & 0.92 & {\bf0.65} & {\bf0.85} & {\bf0.82} \\
      & \text{Augmentation} & 0.56 & 0.97 & 0.96 & 0.55 & 0.62 & 0.58 & - & - & - \\
     \hline

    \end{tabular}
  \caption{Fairness and Accuracy on Adult Income Dataset.}
  \label{table:table1}
 \end{table*}
\end{center}
\section{Experiments}

\subsection{Experimental Setup}
\RaggedRight
This section describes the experiment design and performance of our proposed methodology when
tested on two open-source datasets. Our focus here is on the improved fairness when the synthetically generated dataset is used to train standard prediction algorithms. We first split the dataset into train set (80\%) and test set (20\%), and then pre-process the train data using two pre-processing based bias mitigation techniques. De-biased data coming after the transformation applied in preprocessing step is given as input to various GANs. Then classification model is trained on generated synthetic data and tested on real test dataset which has not been touched and used it for evaluating the model. Data pipeline is shown in Figure \ref{fig:my_image}.



\subsection{Pre-processing based bias mitigation techniques}
\RaggedRight
Pre-processing techniques try to transform the data so that the underlying discrimination is mitigated. We have used two pre-processing techniques:

\subsubsection{K\% Removal Technique}
\RaggedRight
K\% removal is a pre-processing technique which identifies and removes bias inducing samples. This method outlines that the samples with similar features but different output labels and protected attributes are the samples which induces bias. As the protected attributes are not used for modelling, so the samples with similar features but different labels confuses the model and act as a psuedo label noise. To identify bias inducing samples, two groups are formed – one with privileged protected attribute and favourable outcome and other with unprivileged protected attribute and unfavourable outcome. Similar instances between the 2 groups are extracted by calculating cosine similarity of each instance from one group to each instance of other group. Instances with high similarity match are flagged as similar and ranked as per the score. Further, the top 1\%, 2\% or 3\% instances (K is tuneable hyperparameter) from each group which are most biased are removed to preprocess the dataset.


\subsubsection{Data Augmentation Technique }
\RaggedRight

In this method, for every data sample, a new sample is created having the same features and label as that of original sample but with the opposite protected attribute value. Its intuition is to remove the correlation between the protected attribute and the label, and hence remove bias from the model. So, after augmenting synthetic samples, the data will become more balanced and fair. The synthetic points along with original dataset constitute ideal dataset, as label no longer depends on protected attribute, and model is trained on this overall dataset which represents a more equitable world.

Authors have added a subset of those synthetic points which are most realistic and at the same time remove bias. And, to sort the synthetic points in order of most to least realistic data, k-means clustering algorithm is used on raw data such that a set of cluster centers are defined for every protected attribute value(S) and label value(Y), and the inverse of the maximum distance to any cluster center for a point in the synthetic dataset having the same S and Y values is used as a score of how realistic the point is, and then add k-\% of these sorted synthetic points to the real data to create a k-augmented dataset.




\subsection{GANs}

We have evaluated our pipeline on three popular GAN architectures : CTGAN \cite{CTGAN}, Copula GAN \cite{CopulaGAN}, Gaussian Copula GAN \cite{GaussianCopula}. We compared the results on raw data, our proposed pipeline using K\% removal preprocessing technique and Data Augmentation technique.

\subsection{Datasets}
\RaggedRight
We have performed experiments on two datasets – Adult Income Dataset \cite{misc_adult_2} and German Credit Dataset \cite{German}. In Adult Income dataset, task is to predict whether an individual’s income is greater than 50K or less than 50K. This dataset contains two protected attributes (race and gender), standard fairness metrics are evaluated for both features individually. Further, intersectional bias testing is also implemented to comprehensively comment on the fairness aspect. In German Credit Dataset, task is to predict whether a customer should receive a loan or not depending upon the risk. It contains only one protected attribute i.e. gender, standard fairness metrics are evaluated.

\subsection{Evaluation Measures}
\RaggedRight

For measuring performance of the model, we have used Balanced Classification Accuracy (BCA) of the model which is the average of proportion corrects for each class.To measure model bias, we have used Demographic Parity Ratio (DPR) and Equalized Odds Ratio (EOddR) as they are mostly used in literature, and both true positives and false positives are important in our case.

Most of the existing work deals with different attributes individually and missed the important nuisance of intersectionality. There may be chances that the model fair when we consider both protected attributes individually but highly unfair when considered both of them together. That’s why it’s very important to do intersectional bias testing as well where we create subgroups from the combination of multiple protected attributes as biases can often be amplified in those subgroups.We have used following formula to measure intersectional bias and the corresponding results are mentioned in Table \ref{table:table1}. 



\begin{equation}
DPR = \frac{\min\{P(\hat{Y}|A \in sg_i) \forall i \in N \}}{\max\{P(\hat{Y}|A \in sg_i) \forall i \in N \}} 
\end{equation}

\begin{equation}
\small
EOddR = \frac{\min\{P(\hat{Y} = 1|A \in sg_i, Y = y) \forall i \in N, y \in \{0,1\} \}}{\max\{P(\hat{Y}=1|A \in sg_i, Y = y) \forall i \in N, y \in \{0,1\} \}}
\end{equation}
\normalsize


Idea is to measure the value of given fairness metrics for every subgroup(sg) and then take the ratio of minimum and maximum values and as the ratio becomes closer to 1, the lesser will be the discrimination between subgroups.

\begin{table}
\begin{tabular}{|c| c| c| c| c|}
\hline
\multirow{3}{*}{GANs}  &  Preprocessing& \multicolumn{3}{|c|}{\multirow{2}{*}{\textbf{Gender}}} \\
& bias mitigation & \multicolumn{3}{|c|}{} \\
\cline{3-5}
& technique & BCA & DPR & EOddR\\ [0.5ex] 
\hline
  \multirow{5}{*}{CTGAN}& Raw & 0.53 & 0.94 & 0.92 \\
  & \text{1 \% removal} & 0.52 & 0.97 & 0.99 \\
  & \text{2 \% removal} & {\bf0.56} & {\bf0.98} & {\bf0.99} \\
  & \text{3 \% removal} & 0.54 & 0.95 & 0.91 \\
  & \text{Augmentation} & 0.53 & 0.96 & 0.93 \\
 \hline
 
  & Raw & 0.53 & 0.95  & 0.97 \\
  & \text{1 \% removal} & 0.50 & 0.98 & 0.99 \\
  Copula & \text{2 \% removal} & 0.49 & 0.95 & 0.94 \\
  GAN & \text{3 \% removal} & {\bf0.58} & {\bf0.99} & {\bf0.99} \\
  & \text{Augmentation} & 0.56 & 0.95 & 0.92 \\
 \hline
 
  & Raw & 0.53 & 0.99 & 0.99 \\
  & \text{1 \% removal} & 0.48 & 0.85 & 0.88 \\
  Gaussian & \text{2 \% removal} & 0.53 & 0.85 & 0.86 \\
  Copula & \text{3 \% removal} & {\bf0.54} & {\bf0.99} & {\bf1.00} \\
  & \text{Augmentation} & 0.49 & 0.79 & 0.78 \\
 \hline

\end{tabular}
 \caption{Fairness and Accuracy on German Credit Dataset}
 \label{table:table2}

\end{table}

\section{Results}
\RaggedRight
For Adult Income Dataset as shown in Table \ref{table:table1}, we observed that K\% removal is giving fairer results across all three bias tests as compared to the raw data. Also, it is beating state-of-the-art augmentation technique w.r.t gender and race bias tests independent of GAN architecture. We have not performed the intersectional bias testing on Augmentation pre-processing based bias mitigation technique as the authors did not specify how to use their technique for non-binary subgroups classification. For some cases our pipeline has led to improvement in balanced classification accuracy.

For German Credit Dataset as show in Table \ref{table:table2}, We observed that K\% removal has again led to fairer model as compared to the raw data as well as augmentation technique. And, it further increased the balanced accuracy as well.

\section{Conclusion}
\RaggedRight

From the above experiments, we showed that with the proposed pipeline, the generated synthetic data is fair and unbiased independent of the GAN architectures. The above results validates that by removing bias inducing samples from the training dataset, GANs could not amplify the bias unlike the raw data and thus generated fairer synthetic dataset. Further, the results using Data Augmentation technique where biased samples are added is performing worse than the raw data in most cases further validates the argument.

In this paper, we proposed a pipeline to generate fair synthetic dataset. It overcomes the limitations of the existing methods by generating fair synthetic dataset independent of any GAN architectures. Further, the proposed pipeline works seamlessly across the architectures without adding any computationally expensive load.  Finally, promising results on the 2 open-source datasets makes our research well grounded.

\bibliography{example_paper}
\bibliographystyle{icml2022}

\newpage
\appendix
\onecolumn


\end{document}